\documentclass[pageno]{jpaper}

\usepackage[normalem]{ulem}
\usepackage{tikz}
\usepackage{amsmath}
\usepackage{filecontents}
\usepackage{booktabs}
\usepackage{xspace}
\usepackage[noend]{algpseudocode}
\usepackage{algorithm}
\usepackage{listings}
\usepackage{listing}
\usepackage{cleveref}
\usepackage{multirow}
\usepackage{amsthm}
\usepackage{amssymb}
\usepackage{enumitem}

\newcommand{\cred}{\textcolor{red}}

\definecolor{darkgreen}{rgb}{0,0.5,0}

\definecolor{purple}{rgb}{0.3,0.0,0.6}

\definecolor{auburn}{rgb}{0.43,0.21,0.1}

\definecolor{gray}{rgb}{0.7,0.7,0.7}

\definecolor{dgray}{rgb}{0.5,0.5,0.5}

\newcommand{\para}[1]{\noindent \textbf{#1}}

\newcommand{\chkz}{\textcolor{brown}}
\newcommand{\gop}{gOperator\xspace}
\newcommand{\gops}{gOperators\xspace}
\newcommand{\gir}{GIR\xspace}
\newcommand{\ggraph}{\gir-Graph\xspace}
\newcommand{\ggraphs}{\gir-Graphs\xspace}

\newcommand{\sys}{\textproc{PowerFusion}\xspace}

\begin{document}
\title{\sys{}: A Tensor Compiler with Explicit Data Movement Description and Instruction-level Graph IR}
\date{}
\author{ \\ Zixuan~Ma, Haojie~Wang, Jingze~Xing, Liyan~Zheng, Chen~Zhang, Huanqi~Cao \\ Kezhao~Huang, Shizhi~Tang, Penghan~Wang and Jidong~Zhai \\ \\ Tsinghua University}
\thispagestyle{empty}
\maketitle

\begin{abstract}

Deep neural networks (DNNs) are of critical use in different domains.
To accelerate DNN computation, tensor compilers are proposed to generate efficient code on different domain-specific accelerators.
Existing tensor compilers mainly focus on optimizing computation efficiency.
However, memory access is becoming a key performance bottleneck because the computational performance of accelerators is increasing much faster than memory performance.
The lack of direct description of memory access and data dependence in current tensor compilers' intermediate representation (IR) brings significant challenges to generate memory-efficient code.

In this paper, we propose \sys{}, a tensor compiler that can generate high-performance code for memory-intensive operators by considering both computation and data movement optimizations.
\sys{} represent a DNN program using \gir, which includes primitives indicating its computation, data movement, and parallel strategies.
This information will be further composed as an instruction-level dataflow graph to perform holistic optimizations by searching different memory access patterns and computation operations, and generating memory-efficient code on different hardware.
We evaluate \sys on NVIDIA GPU, AMD GPU, and Cambricon MLU, showing speedup up to $1.97\times$, $2.93\times$, and $16.91\times$ ($1.28\times$, $1.23\times$, and $2.31\times$ on average), respectively, compared to current most performant frameworks.




\end{abstract}


\section{Introduction}
\label{sec:intro}

Deep neural networks (DNNs) have been widely used in a number of important domains, such as computer vision (CV), natural language processing (NLP), and so on. Due to the massive requirement of DNN models for computation power, domain-specific accelerators have been developed to improve DNN efficiency. 
While the computational performance of accelerators has rapidly increased in recent years, memory performance is lagging far behind.

\begin{figure}[t]
    \centering
    \includegraphics[width=0.9\linewidth]{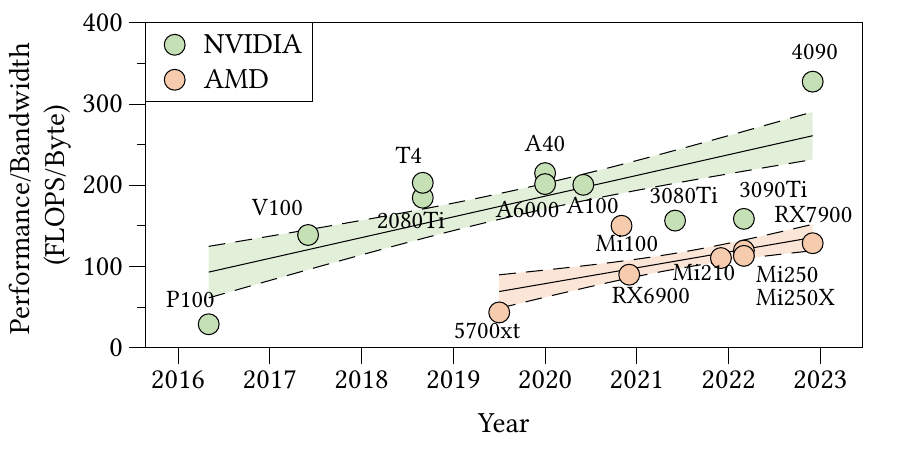}
    \caption{Performance trend of computation and memory performance in recent years.}
    \label{fig:trend}
\end{figure}

\Cref{fig:trend} illustrates the trend of half-precision performance and memory bandwidth of typical accelerators from $21.2$ TFLOPS and $732$ GB/s (NVIDIA Tesla P100) in 2016 to $330.3$ TFLOPS and $1,008$ GB/s (NVIDIA RTX 4090) in 2023. During this period, the ratio of computation to memory performance has increased by $11.3 \times$, making memory performance the main bottleneck of DNN models.
Moreover, the divergence in architectures and memory hierarchies of different accelerators also brings significant optimization challenges to designing performance portable DNN systems.
Therefore, optimizing memory efficiency is becoming essential to fully explore hardware performance.

Tensor compilers are designed to generate efficient code on different accelerators.
There are two main approaches to generating memory-efficient code: improving hardware memory bandwidth utilization and minimizing low-level (large and slow memory hierarchy) memory access. 
The former requires the program's memory access pattern to align with the hardware characteristics to achieve peak performance, while the latter involves analyzing data reuse relationships and reusing data on high-level memory to reduce low-level memory access.
Achieving both requires the tensor compiler to optimize memory access directly, which requires an \textbf{explicit description of data movement} in their intermediate representation (IR). 
Additionally, the compiler also needs to analyze \textbf{data dependencies at a fine granularity} to reuse data using high-level memory hierarchy, improving overall memory performance.




Existing tensor compiler faces significant challenges in meeting these requirements. For example, TVM~\cite{tvm} and Ansor~\cite{ansor}represent a tensor program with an abstraction of compute and schedule, inspired by Halide~\cite{halide}. 
These compilers first convert DNN models into a loop-based IR (\textit{compute}) and apply a series of optimizations (\textit{schedule}) to transform DNN programs to find better performance. 
Consequently, they generate efficient code tailored to specific architectures. However, memory-intensive programs with this abstraction incur three main challenges: 


\para{Implicit data movement representation} In TVM, the pattern of data movement is implicitly represented through a schedule, making it challenging to assess memory performance during optimization and optimize memory performance effectively. 

\para{Schedule search order.} When merging multiple computing operations, TVM first fuses multiple loops into one and then applies schedules to generate efficient code. This generation order prevents the application of different memory access patterns from distinct computation operations, missing potential optimization opportunities. 

\para{Coarse-grained dependence analysis} 
Existing tensor compilers represent DNN programs as perfectly nested loops, which limits dependence analysis only to loops.
This dependence analysis granularity does not align with the data's granularity in the memory hierarchy, leading to missing opportunities to reduce memory access through data multiplexing. 

These constraints make it challenging for tensor compilers to achieve optimal memory performance. 
To address these limitations, we propose \sys{}, a tensor compiler that considers both computation and data movement patterns to address the challenge of generating memory-efficient code for various architectures. 
\sys{} represents tensor programs using \emph{\gir}, which describes operators using three primitives, parallel, computation, and data movement.
The parallel primitive indicates parallel strategies for a given computation. 
The computation and data movement primitives, which we call \emph{\gops}, are constructed as nodes of an instruction-level dataflow graph, called \emph{\ggraph}, and the edge between them represents the data on certain memory hierarchy and also their dependence. 
By explicitly expressing memory access operations and instruction-level dependence, 
\sys{} can search different memory access patterns for multiple computing operations, thus supporting a more thorough optimization space for memory-intensive operators.

With this representation, \sys{} can generate different \ggraphs for each computation with various memory access patterns to find the configuration that maximizes memory bandwidth utilization. By explicitly representing data dependence at the instruction level, \sys{} can analyze fine-grained dependence and apply graph optimization to reduce redundant memory access and reuse data in the proper memory hierarchy with the lowest cost. This optimization is automatic and unlimited, requiring no rules to guide the merging process and supporting more complex tensor programs. Additionally, we provide a performance model in \sys{} to evaluate the key performance indicators of \ggraph. This model can assist in deciding which operators to generate or use from a DNN library and which \ggraph performs best and is worth generating. Finally, the methods employed by \sys{} are cross-platform with platform-independent optimization strategies. We have designed a general abstraction that can accommodate various architectures with different parallel structures and memory hierarchies. By configuring the \ggraph generation and searching process, \sys{} can generate optimized code for a specific hardware platform, requiring developers  to 
 only implement instruction-to-instruction mappings for computation and data movement to adapt \sys{} to a new hardware platform.

We have implemented \sys{} from scratch and support multiple architectures, including NVIDIA GPU, AMD GPU, and Cambricon MLU. Our evaluation with various DNN models demonstrates that \sys{} can efficiently generate code on different platforms, achieving higher performance than native computing systems like TensorRT~\cite{tensorrt} on NVIDIA GPU and MagicMind~\cite{magicmind} on Cambricon with $1.28\times$ and $2.3\times$ respectively. By using \gir descriptions, \sys{} can seamlessly adapt to different architectures, requiring only up to $1,000$ lines of code for adaptation.

This paper makes the following contributions:

\begin{itemize}

\item We present \gir, which targets optimizing memory performance by explicitly representing data movement patterns and fine-grained data dependence through instruction-level graph descriptions.

\item We design a set of optimization strategies that include searching and transformations on \gir, to reduce memory access and improve memory efficiency.

\item We propose a hardware abstraction and develop an end-to-end tensor compiler that utilizes this abstraction, and generating optimized code for various architectures.

\item \sys{} is implemented on different hardware, including NVIDIA GPU, AMD GPU, and Cambricon MLU, and achieves up to $1.97\times$, $2.93\times$, and $16.91\times$ ($1.28\times$, $1.23\times$, and $2.31\times$ on average), respectively, compared with the most efficient DNN framework on each hardware.

\end{itemize}

\begin{figure*}[t]
    \centering
    \includegraphics[width=0.9\textwidth]{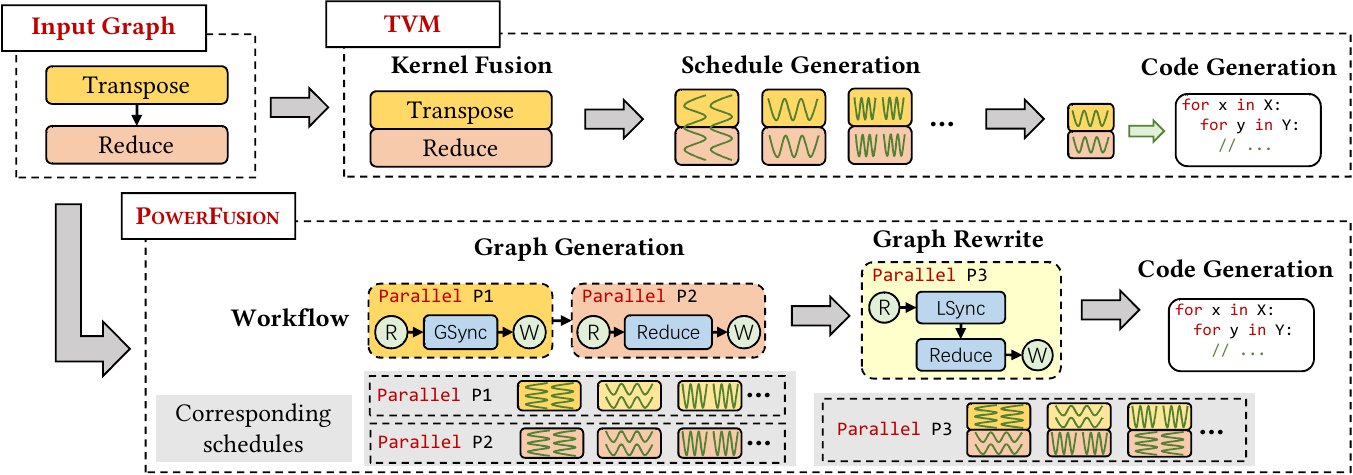}
    \caption{Illustration of TVM and \sys{}'s approach.}
    \label{fig:tvm}
\end{figure*}

\section{Background}
\label{sec:back}

TVM and Ansor adopt the compute/schedule separation idea and use auto-tuning techniques to search over different schedules for code generation, working with loop-based IR. As illustrated in \Cref{fig:tvm}, when optimizing multiple computation operations, TVM first merges the loops of different operations and then searches for the execution plan of the merged loop before generating codes. However, this approach leads to two limitations. Firstly, during dependence analysis, it becomes impossible to analyze the reuse relationship of the data blocks that correspond to each data movement operation on the high memory hierarchy, preventing many operations from being fused. Secondly, the merging-first-then-scheduling approach makes it impossible for different operators to adopt different execution plans, which results in suboptimal in-memory performance.

Polyhedral works, represented by PPCG~\cite{ppcg} and PLUTO~\cite{pluto}, are capable of analyzing element-level dependence in a loop-based program, by representing the program mathematically as points in a high-dimensional integer space. The program can then be optimized by transforming the space. This type of work typically employs integer linear programming to find a mathematically optimal transformation. However, due to the lack of consideration for specific hardware constraints, the optimized program may not perform well when running on real hardware. Considering the huge overhead of integer linear programming, it is also hard to adopt auto-tuning techniques.

Another approach, represented by TensorFlow-XLA~\cite{tensorflow_xla} and DNNFusion~\cite{dnnfusion}, uses kernel fusion at the computational graph level to reduce memory access. It generates the fusion plan on the computation graph and then employs the code generation tool to generate device code. TensorRT~\cite{tensorrt} is another type of work that uses a series of rules to map complex computation operations in the DNN model to manually-optimized kernels. This approach takes the operation as the granularity of analysis, and its extensibility is limited by the capability of the back-end operator library or code generation.

\begin{figure*}[t]
    \centering
    \includegraphics[width=0.95\textwidth]{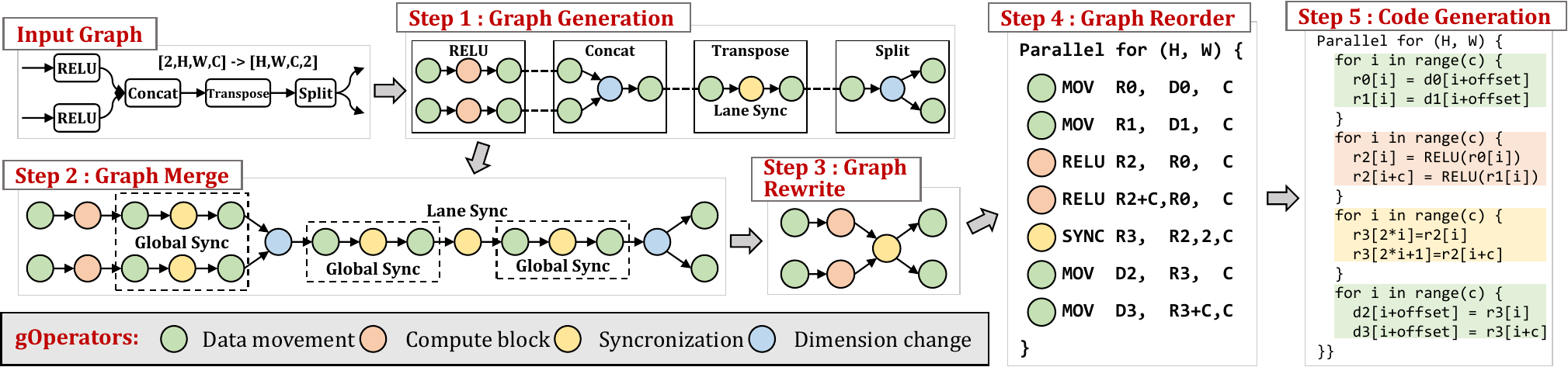}
    \caption{Overview and motivating example for \sys{}.}
    \label{fig:example}
\end{figure*}

To address these limitations, \sys{} utilizes instruction block and memory slice as the granularity of analysis. This granularity matches the computation and data movement operations that are actually performed on the hardware, providing the required information for data reuse. By organizing the program as a graph with computation and data movement operations at this granularity, the dependence analysis overhead is significantly reduced. As shown in \Cref{fig:tvm}, compared to TVM, \sys{} first searches for the execution plan of different operations and then combines different execution plans through fine-grained dependence analysis to achieve higher memory performance.

Various domain-specific accelerators, such as NVIDIA GPUs and Cambricon MLUs, generally employ a hierarchical memory structure based on scratch-pad memory, comprising several layers such as register file, shared memory, and global memory. This complex and varied memory hierarchy presents two main challenges to memory performance optimization for accelerators: 1) the deep coupling of the memory hierarchy and parallel structure, and difficulty in determining the optimal memory hierarchy for data usage. 2) complex data movement between different memory hierarchies, and difficult to optimize.

To address these challenges, we abstract the memory hierarchies of different hardware in a unified abstraction and represent the characteristics of different hardware in its attributes. This approach enables us to provide unified optimization and simplify memory performance optimization for different architectures. Furthermore, by incorporating synchronization operations in GIR, we can efficiently analyze the highest memory level that data can use, allowing \sys to provide optimal memory performance optimization for different architectures.

\section{Overview}
\label{sec:overview}

\Cref{fig:example} provides an overview of \sys{}, along with a running example of how it optimizes memory performance for a real-world DNN model. 
The given model fragment is from ShuffleNet, which merges two tensors into one, shuffles it, and then divides it into two new tensors. 
To optimize this model, \sys{} takes this model fragment as input, and converts it to instruction-level graph IR, called \emph{\gir}.
Then \sys{} optimizes this model by \gir-based optimizations, and finally generates high-performance code. 
The \gir abstraction will be introduced in \Cref{sec:graph}.

\gir-based optimizations take a computation graph as input and apply graph generation to each operator separately to obtain the \ggraph of each operator. 
Next, it searches and merges these \ggraphs to obtain the \ggraph corresponding to the computation graph.
With the merged \ggraph, \sys{} then applies graph rewriting rules to optimize the \ggraph using three graph rewriting rules, resulting in a reduction of memory access amount, significantly improving memory performance. Graph rewriting and \gir-based optimization process will be introduced in \Cref{sec:graph-rewrite} and \Cref{sec:graph-gen}, respectively.

Finally, \sys{} rearranges the optimized \ggraph and sequentially generates code for specific hardware. Code generation techniques will be discussed in \Cref{sec:code-gen}.


\section{\gir Abstraction}
\label{sec:graph}

\begin{figure}[t]
    \centering
    \includegraphics[width=0.45\textwidth]{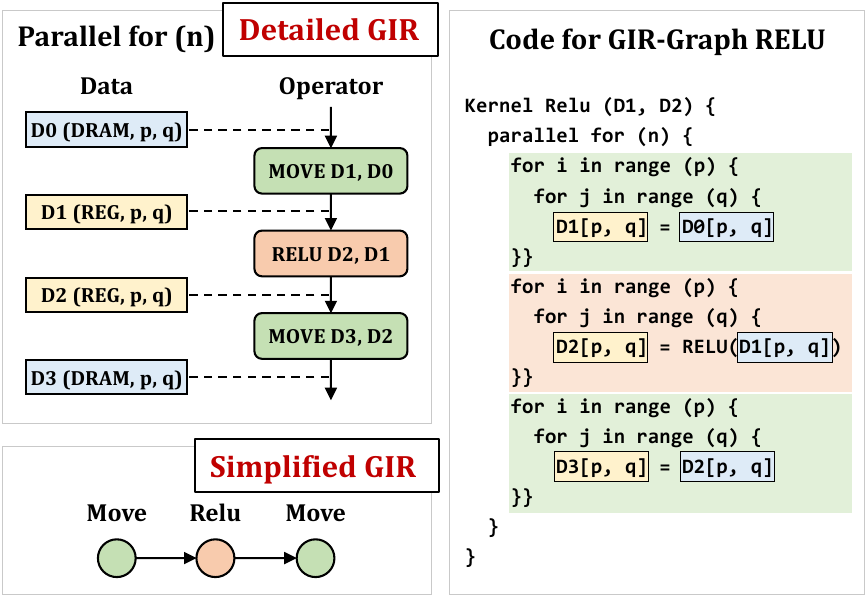}
    \caption{Illustration of \gir abstraction.}
    \label{fig:micro-graph}
\end{figure}

This section presents the design of \gir, which aims to address the lack of a comprehensive description of data movement between memory hierarchies in existing tensor compilers, thereby hindering memory access performance optimization. The core idea of the \gir is to represent a tensor program as a graph consisting of computation and data movement primitives, called \gop{}s, with the granularity of instruction block at a specific level of parallelism. The \ggraph and the device code for a \texttt{RELU} operation as shown on \Cref{fig:micro-graph}.

\gir is represented as a dataflow graph wrapped by parallel primitives. Each node of the \ggraph denotes an instruction-level computation or data movement \gop, and each edge represents a memory slice on a specific memory hierarchy. The parallel primitive indicates the parallel strategies of this instruction-level dataflow graph. This approach offers two advantages: first, all computation and data movement operations are expressed with proper granularity, making them easier to optimize; second, the dependences between computation and data movement are clear and easy to analyze. The graph structure in the intermediate representation also enables the use of graph algorithms to optimize the \ggraph and achieve end-to-end optimization effects.

\paragraph{Memory slice} In \gir, data is represented as a memory slice, which is a set of contiguous elements on a tensor. Each memory slice has four attributes: \textit{memory hierarchy}, \textit{num}, \textit{width}, and \textit{stride}. 
\textit{Memory hierarchy} indicates which level of memory hierarchy this memory slice is stored, e.g., DRAM or SRAM.
The \textit{num} attribute specifies the number of continuous segments, each with a length of \textit{width} elements and a fixed stride of \textit{stride} between segments. 
This format has two advantages. 
First, it aligns with best practices for memory performance since most memory hierarchies, such as DRAM and SRAM, are optimized for continuous access. 
Second, it matches the granularity and shape of computation instruction blocks, which are often designed in this format. The two-dimensional format enables fine-tuning of data movement operations without significant performance loss.

\paragraph{Computation \gops} 
The \gir uses computation abstraction, namely computation \gop, at the instruction block level, where each computation \gop corresponds to a set of instructions corresponding to the optimized computation kernel in the code. 
The computation in \gir is defined by three types of \gop{}s: element-wise, reduce, and broadcast. 
Element-wise \gops perform element-to-element computations, such as arithmetic and activation. 
Reduce and broadcast \gops handle dimension expansion and contraction. 
Note that most wrapped computation can also be represented as the above three \gops, e.g., we can represent matrix multiplication with element-wise and reduce operations.
\sys{} will automatically choose the representation method to achieve better performance (an example is shown in \Cref{fig:graph-gen}).
Using this fine-grained computation primitive, we can describe program computation and optimize their schedules for higher performance without considering hardware-related information such as instruction ordering.

\paragraph{Data movement \gops} 
Data movement \gops in \gir describe the transfer of data within and across different memory hierarchies. 
These operations require the input and output to have the same pattern, which includes the number of elements, the width of each element, and the stride between consecutive elements. 
By explicitly expressing data movement operations, \sys{} enables efficient handling of data transfer and optimization of memory performance.


\gir also explicitly introduces a special case of data movement \gops, called synchronization \gops, to ensure correctness between data movement \gops. 
For instance, in CUDA, synchronization operations within a thread block are necessary when different warps access the same shared memory object. 
The scope attribute of a synchronization \gop in \gir specifies its scope, which determines the implementation of the operation. 
In CUDA, intra-warp synchronization uses warp shuffle, while block synchronization uses shared memory. 
In \gir, a synchronization operation not only implies waiting but also indicates a change in the memory slice pattern, reflecting that different data movement operations use different memory slice patterns to read and write the same memory data. 
By incorporating synchronization operations, \gir can better analyze the dependence between data and ensure program correctness.

\paragraph{Parallel}
In \sys{}, we define the parallel granularity as the smallest memory performance granularity. 
For example, on GPUs, it corresponds to a streaming multiprocessor (SM), which is equivalent to a warp in the CUDA programming model. 
While on CPUs and Cambricon MLUs, it corresponds to a core or IPU, which is equivalent to a thread. We refer to this smallest unit as the parallel unit and assign an index to each unit. 
We then map the nested parallel structure of the hardware onto the index space, with distinct segments contiguously allocated to each structure. 
For instance, in CUDA, the indexes of all warps in the same block are contiguous. 
This approach enables us to analyze the parallel scope of each data movement operation and eliminate the impact of different parallel nesting structures across architectures. 
As a result, we can streamline the code optimization process for more efficient execution.

\section{\ggraph Rewriting}
\label{sec:graph-rewrite}

As discussed in \Cref{sec:graph}, \gir can express data dependence between computation and data movement \gops, enabling additional analysis and optimization opportunities. To optimize a \ggraph, we propose graph rewriting techniques using a set of rules called rewrite rules, which are commonly used for graph transformations. In \sys{}, we propose three types of rewriting rules and an optimization strategy to deploy these rules on \gir.

\subsection{Rewrite Rules}

\begin{figure}[t]
    \centering
    \includegraphics[width=0.45\textwidth]{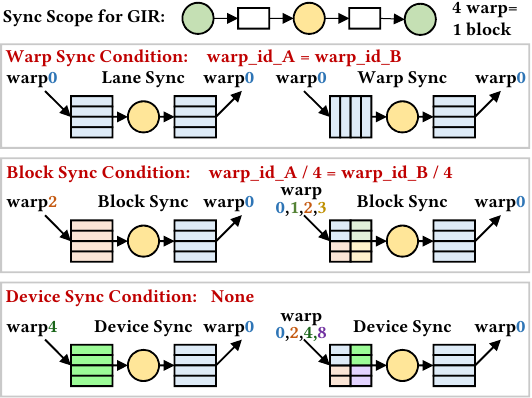}
    \caption{Demonstration of how \sys{} determine synchronization scope.}
    \label{fig: sync}
\end{figure}

\paragraph{Synchronization insertion} The goal of synchronization insertion is to increase the memory hierarchy of memory slices in the \gir and to add synchronization operations to ensure the correctness of the program. When multiple data movement \gops access the same memory slice, synchronization is necessary among the parallel units participating in the \gop. The requirement for synchronization arises from the fact that different parallel units before and after synchronization access the same memory slice. Essentially, the process of read-synchronize-write involves exchanging data between various parallel units.

Different memory hierarchies correspond to different parallel scopes. For instance, on NVIDIA GPUs, data exchange within the same warp is stored in registers and synchronized using warp shuffle. When data is exchanged among different warps within a thread block, shared memory is utilized, and synchronization takes place at the thread block. Therefore, the highest memory hierarchy for data storage can be determined by analyzing the parallel units of data exchange for the read and write operations of the same memory block.

The mapping between the memory hierarchy and parallel units is specific to each hardware architecture and is expressed through a set of conditions. Although these conditions are associated with the kernel execution configuration, their format is independent of the hardware. For example, in a CUDA kernel with four warps per thread block, the synchronization scope condition is shown in Figure \ref{fig: sync}. Each square in the figure represents a memory slice, and its pattern represents its memory slice pattern. The different colors represent different parallel units that access the data in the corresponding memory. The synchronization scope is determined by the units involved in the reading and writing operations. If the unit that writes is the same as the unit that reads, the synchronization scope is restricted to the warp level. In cases where the pattern of reading and writing is the same, the scope of the synchronization \gop is restricted to the lane level, and the operation has no effect. If the result of dividing the IDs of the unit that reads or writes by 4 is the same, then the synchronization scope is restricted to the block level. Otherwise, the scope is restricted to the device level.

Using the conditions mentioned above, we define a synchronization insertion operation. When two consecutive data movement \gops are encountered (synchronization \gop is allowed between them, as is the case for all consecutive \gops in this section), we analyze the synchronization relationship between the two memory slices and increase their memory hierarchy to the highest level. A new synchronization \gop is then inserted into the rewritten graph, with the exception of lane synchronization, which only inserts a new memory slice without synchronization. This process improves the memory performance of the \gir and ensures program correctness.

\paragraph{\gop merging} Although we can improve the memory hierarchy of memory slices by applying synchronization insertion, there are still a large number of redundant data movement \gops in \gir and introduce additional data movement and synchronization overhead, meanwhile generating redundant memory slices, thereby reducing execution performance. This redundancy falls into two main categories: read-after-write and read-after-read. Read-after-write means that two consecutive data movement \gops sequentially operate on the same memory slice, write first, and then read. This kind of data movement may be redundant. Read-after-read means that two data movement \gops read the same memory slice. When there is no synchronization relationship between the two \gops, this combination will cause redundant data movement.

To solve this redundancy problem, we propose the \gop merging rule, which consists of two rules:

\begin{itemize}
\item Replace two consecutive data movement \gops with a new data movement \gop whose input is the input of the first \gop and output is the output of the last \gop. If the memory slice pattern read and written is consistent, it will be replaced by a null \gop.
\item Replace two data movement \gops that read the same memory slice without dependence with one \gop.
\end{itemize}

By combining multiple \gops into one, this rule can reduce memory access for \ggraph thus optimizing the memory performance.

\begin{figure}[t]
    \centering
    \includegraphics[width=0.45\textwidth]{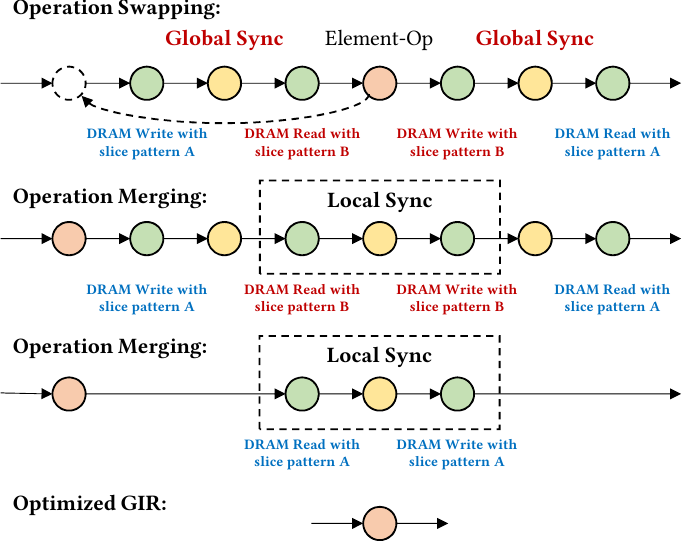}
    \caption{Example of graph rewriting on \gir.}
    \label{fig:graph-trans}
\end{figure}

\paragraph{\gop swapping} Applying these two rules mentioned above does not guarantee optimal memory performance of the program. In fact, a \gir, as shown in \Cref{fig:graph-trans}, cannot be further optimized. This is because the program is constrained by two global synchronization \gops, resulting in the generation of four DRAM read and write operations. This is due to the disconnect between data movement and synchronization \gops, which makes it impossible to apply optimization rules.

To solve this problem, we can move the computation \gops. For example, element-wise \gops represented by arithmetic computations, activation functions, etc., are interchangeable with data movement, synchronization, broadcast, and some of reduce \gops. This interchangeability allows us to move element-wise \gops, providing more opportunities for optimization. \Cref{fig:graph-trans} shows a process of optimizing \ggraph through \gop swapping. By applying \gop swapping, the \gir can further apply \gop merging rules, thereby reducing the total execution time of the program.

\subsection{Optimization Strategy}

To optimize the memory usage of a program, we can apply three graph rewriting rules on the \gir. These rules ensure that the program's theoretical memory performance does not decrease after applying the rules. Therefore, we can apply as many rewriting rules as possible during the optimization process. However, if all the rules are irreversible, we can use a greedy strategy instead of the search-based method to quickly find the optimal solution. Among the three rules, synchronization insertion and \gop merging are irreversible and can only be applied in one direction. On the other hand, the \gop swapping rule is reversible, so our optimization approach only allows forward swapping of element-wise \gops to make it irreversible.

With the revised rules, we propose a greedy optimization strategy outlined in the pseudo-code shown in \Cref{algo:inter-expr}. The strategy applies synchronization insertion and \gop swapping as much as possible, followed by \gop merging until \gop merging can no longer be applied. Since the rules are irreversible, the optimization is directional and guaranteed to terminate. This algorithm enables us to quickly obtain an optimized \ggraph, and in practice, the solution time is negligible. Thus, we can search complex graph structures and experiment with different combinations of rules to optimize \ggraph in various applications. This fast optimization method is a valuable tool for optimizing tensor programs with \gir.

\begin{algorithm}[t]
\caption{\ggraph optimization algorithm.}
\label{algo:inter-expr}
\small
\begin{algorithmic}[1]
\State {\bf Input:} An input \gir $G$
\State {\bf Output:} The optimized \gir $G$
\State
\Function {InsertSync}{$G$, $n_0$, $n_1$}
    \If {$n_0$ and $n_1$ are consecutive}
        \State Add sync \gop between $n_0$ and $n_1$
    \EndIf
\EndFunction
\State
\Function {MergeOp}{$G$, $n_0$, $n_1$}
    \If {$n_0$, $n_1$ is mergable}
        \State merge $n_0$, $n_1$ and insert merged node to $G$
        \State \Return true
    \Else
        \State \Return false
    \EndIf
\EndFunction
\State
\Function {SwapOp}{$G$, $n_0$, $n_1$}
    \If {$n_0$, $n_1$ is swappable}
        \State swap $n_0$ and $n_1$
    \EndIf
\EndFunction
\State
\While {true}
    \For {$n_0$, $n_1$ in $G$.nodes}
        \State \Call{InsertSync}{$G$, $n_0$, $n_1$}
    \EndFor
    \For {$n_0$, $n_1$ in $G$.nodes}
        \State \Call{SwapOp}{$G$, $n_0$, $n_1$}
    \EndFor
    \State flag = false
    \For {$n_0$, $n_1$ in $G$.nodes}
        \If {\Call{MergeOp}{$G$, $n_0$, $n_1$}}
            \State flag = true
        \EndIf
    \EndFor
    \If {flag == false}
        \State break
    \EndIf
\EndWhile
\State 
\State \Return $G$
\end{algorithmic}
\end{algorithm}

\section{\ggraph Generation}
\label{sec:graph-gen}

The primary objective of \sys{} is to optimize the memory performance of tensor programs from end to end. To achieve this, we generate \ggraphs for the input model, which can be further optimized using Graph Rewriting techniques and generate efficient code. In this section, we will explain how the \ggraphs are generated from the model.

\begin{figure}[t]
    \centering
    \includegraphics[width=0.45\textwidth]{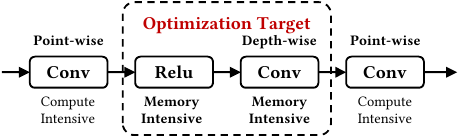}
    \caption{Example of a model fregment in EfficientNet.}
    \label{fig:subgraph}
\end{figure}

\paragraph{Preprocessing of computation graph}
The input to \sys{} is a model represented as a computation graph. 
As the primary optimization goal of \sys{} is memory performance, operators bounded by computing performance directly use DNN libraries like cuDNN on NVIDIA GPU or CNNL on Cambrian MLU. 
We determine whether to call the DNN library based on the operators' computation and memory access amount, rather than its type. 
As a typical example shown in \Cref{fig:subgraph}, which is a subgraph in the EfficientNet model, the depth-wise convolution operators are memory-intensive.
Thus it will be expressed using \gir and jointly optimized with its preceding operator \texttt{RELU}.
On the other hand, the two point-wise convolutions are computation-intensive, so \sys{} chooses to call the DNN library directly.
\sys{} will also adjust the data layout of the operators when calling the library function to achieve better performance by introducing additional transpose operators.
These transpose operators will be jointly optimized with other memory-intensive operators to further improve the performance.
After these optimizations, the model is split into separated subgraphs containing only memory-intensive operators. 

\begin{figure}[t]
    \centering
    \includegraphics[width=0.45\textwidth]{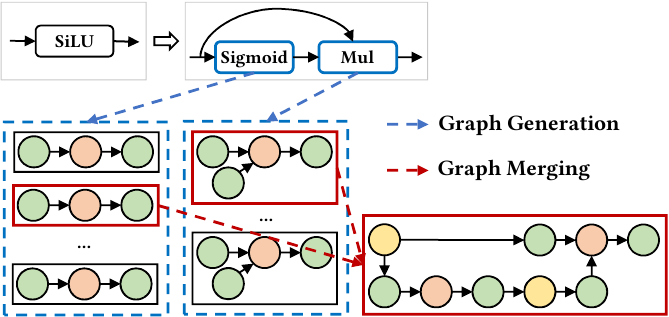}
    \caption{Illustration of \ggraph generation and merging.}
    \label{fig:graph-gen}
\end{figure}

\paragraph{\ggraph generation} 
The next step in \sys{} is to generate \ggraphs from the subgraphs composed of memory-intensive operators. 
As described in \Cref{sec:back}, searching the execution plan of different operators and then applying the combination can effectively improve the search space of the program, leading to higher performance. 
To achieve this, complex operators need to be split. 
For instance, the \texttt{SILU} operator is a composite operator comprising a \texttt{SIGMOID} and a \texttt{MUL} operator. 
\sys{} splits all such composite operators into four basic operators: element-wise, broadcast, reduce, and transpose. Note that these basic operators are high-level operators on the computation graphs, not the same as \gir's \gop{}.
These operators can apply predefined code templates to generate \ggraphs.

For each basic operator, we define several corresponding \ggraph structures. 
We take the \texttt{REDUCE} operator as an example. 
The reduction with one parallel unit and with multiple parallel units requires different computation and data movement \gop{}s that correspond to different graph structures. 
Besides, with a fixed graph structure, each operator has several implementation methods, including parameters such as parallelism at execution time and the tiling size for specific implementations. 
\sys{} traverses the possible values of these parameters on the graph structure. 
The combination of the graph structure and the parameters are defined as a \gir template that can generate a set of \ggraphs for a specific basic operator. 
The graph generation of the \texttt{SIGMOID} and the \texttt{ADD}, both of which generate multiple \ggraphs, as shown in \Cref{fig:graph-gen}.

\paragraph{\ggraph merging}
After generating \ggraphs for each basic operator, \sys{} attempts to merge several \ggraphs to a larger one, thus joint optimizing multiple operators. 
Our design allows \ggraph merging only if the parallel structures of the two \ggraphs match and there are no external dependencies between the operators on the corresponding computation graph. 
For instance, two operators can be independent or directly connected. 
Given this rule, we can efficiently generate a merging plan when we choose a \ggraph for each basic operator.

When merging two \ggraphs, we also insert global synchronization operations between data movement \gop{}s that read and write the same tensor. For sequential reads and writes, a global synchronization \gop{} needs to be inserted between two data movement \gop{}s. For parallel reads, a global synchronization operation needs to be inserted before the two \gop{}s. The resulting merged \ggraph can be rapidly optimized to achieve high memory performance by applying Graph Rewriting techniques. \Cref{fig:graph-gen} shows an example of how synchronization operation is inserted into merged \ggraphs.

To explore the search space as much as possible, \sys{} traverses all \ggraphs of all basic operators and generates all possible merge plans. However, the direct search approach for this step is computationally expensive. To address this issue, we introduce a performance model. Since all \gop{}s in a \ggraph are regular and performance-independent, we can accurately predict the running time of the \ggraph using the performance model. We can then prune the search space and efficiently obtain an optimization plan that guarantees the optimization result.

\section{\gir Code Generation}
\label{sec:code-gen}

The design principle of \sys{} involves abstracting different hardware into a unified structure and extracting hardware features as parameters. This abstraction allows us to optimize programs and generate code for parallel computation and data movement \gop{}s that are specific to the hardware being used. By decoupling hardware and optimization, the system becomes more extensible.

Before generating device code, \sys{} reorders the \ggraph to determine the execution order of computation and data movement \gop{}s. This determines the reuse of memory slices in the physical memory hierarchy. The reordering process is similar to traditional compiler instruction reordering and can use existing methods for optimization. After determining the execution sequence, \sys{} generates \ggraphs as code in the native programming language and compiles it into the device executable using native compiler.

During the code generation phase, \sys{} generates parallel structure code related to hardware information first, and then sequentially generates device codes for computation and data movement \gop{}s in order. The generated code is compiled using a native compiler and evaluated to ensure optimal performance. This design enables the low porting cost of \sys{} to new hardware, as porting only requires mapping the parallel abstraction and primitives for computation and data movement. For example, porting to NVIDIA GPU and AMD GPU requires less than $1,000$ lines of new code for each. We will explain how \sys{} adapts to different platforms and generates efficient code using NVIDIA GPU and Cambrian as examples.

\paragraph{NVIDIA GPU}

The GPU is the most commonly used accelerator in AI domain, with vendors such as NVIDIA and AMD having similar hardware architectures and programming models. To illustrate how \sys{} is ported to GPUs, we show how it maps to CUDA on NVIDIA GPUs.

CUDA's parallel structure consists of three layers: thread block, warp, and thread. To map the parallel unit of \gir onto CUDA, we map it to the warp, as described in \Cref{sec:graph}. The warp is the smallest scheduling unit in CUDA with independent performance. In contrast, threads in CUDA have dependent memory access and scheduling and correspond to SIMD lanes in the SIMD architecture, which makes them unsuitable as parallel unit in \gir. During the search process, the number and size of thread blocks in CUDA can be configured and used as search parameters. 

For the instructions of computation and memory access operations, the CUDA program represents the program executed on a specific thread. Thus, in \gir, an implementation that says "read contiguous memory containing $n \times 32$ elements" would generate "read $n$ element with stride $32$".

\begin{figure}[t]
    \centering
    \includegraphics[width=0.4\textwidth]{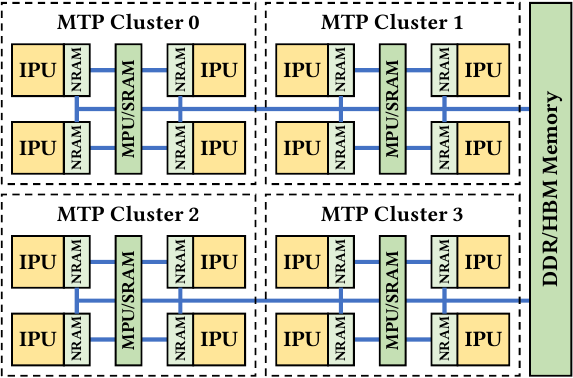}
    \caption{The architecture of the Cambricon MLU.}
    \label{fig:cambricon-arch}
\end{figure}

\paragraph{Cambricon MLU} Cambricon MLU is a domain-specific architecture that differs significantly from GPUs in terms of architecture. It has two layers of scratch-pad memory, NRAM and SRAM, which are located in the scope of one IPU or one MTL Cluster, respectively. However, data movement operations on the MLU platform require direct memory access (DMA), which can introduce overhead and limit the memory performance achievable by synchronously performing computation and data movement operations on each IPU.

To improve memory efficiency, \sys{} applies pipeline parallelism, a common optimization technique on the MLU platform. In this technique, multiple pipelines are executed on a single IPU simultaneously, enabling data movement to overlap with computation. During code generation, \sys{} selects the parallel unit as a pipeline on the IPU and interleaves multiple pipelines to generate execution code. Synchronization operations are then inserted to ensure correct execution. This pipeline parallelism optimization introduces a new parallel dimension under the MTP Cluster and IPU, and does not affect any existing optimization on \gir.
The successful porting of \sys{} to Cambricon MLU platform demonstrates its strong extensibility.

\section{Evaluation}
\label{sec:eval}

\begin{figure*}[t]
    \centering
    \includegraphics[width=1\textwidth]{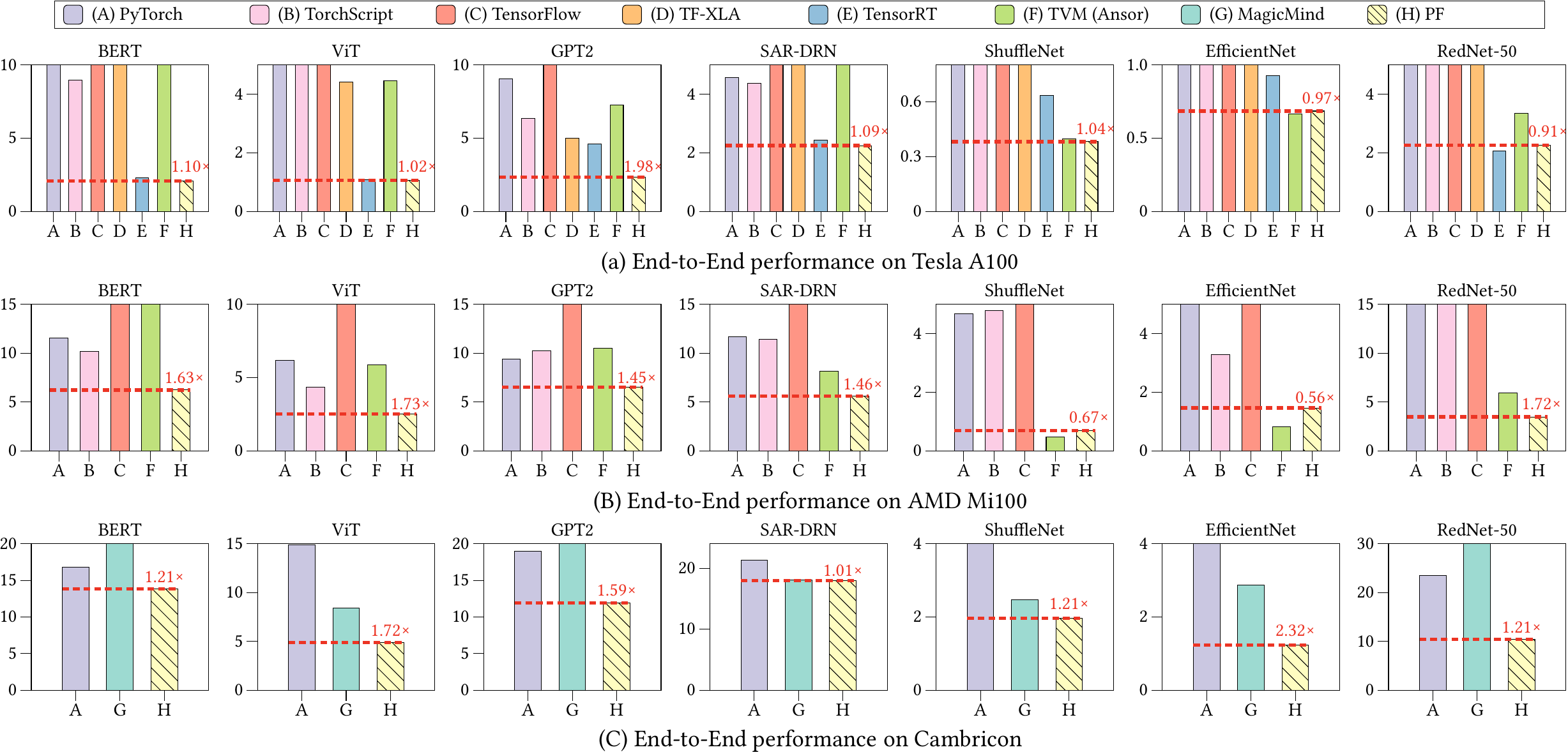}
    \caption{End-to-end performance.}
    \label{fig:e2e}
\end{figure*}



\subsection{Evaluation Setup}

We evaluated \sys{} on three different hardware platforms: NVIDIA GPU, AMD GPU, and Cambricon.
For NVIDIA GPU, we used a Tesla A100 40GB PCIe GPU, which supports Tensor Core for accelerating matrix computation. The peak performance of Tesla A100 on TF32 datatype is $156$ TFLOPS, and its peak DRAM bandwidth is $1,555$ GB/s. We used CUDA version 11.7.0 for the evaluation and set the memory and application clocks to their maximum values.

For AMD GPU, we used INSTINCT MI100, which is optimized for high-performance matrix computation. The peak performance of MI100 on matrix core is $46.1$ TFLOPS, and its peak DRAM bandwidth is $1,200$ GB/s. We used ROCm version 4.3 for the evaluation.

For Cambricon, we used MLU-370x4, which has a peak FP32 performance of $24$ TFLOPS and peak DRAM bandwidth of $307.2$ GB/s. We used BANGC version 1.0 for the evaluation.

These hardware platforms were chosen to represent a diverse range of architectures and to provide a comprehensive evaluation of \sys{}'s performance.

\subsection{End-to-End performance}

We evaluated seven models on \sys{} to compare with different baselines on various architectures.

\paragraph{DNN Models}
We evaluated \sys{} on seven models that include three transformer models and four CNN models for model diversity.
BERT~\cite{bert} and GPT-2~\cite{radford2019language} are popular language models that use transformer architecture for natural language processing tasks. The difference is that GPT-2 is autoregressive that should run the model as many times as the number of tokens to generate.
Vision Transformer (ViT)~\cite{vit} applies transformer model to computer vision tasks such as image classification and object detection.
SAR-DRN~\cite{sardrn} and EfficientNet~\cite{efficient_net} is a popular CNN models for super-resolution image generation;
ShuffleNet~\cite{shuffle_net} and RedNet-50~\cite{li2021involution} introduce more complex memory-intensive operators to CNN models.
In the evaluation, we set batch size of all models to $1$ except GPT-2, whose batch size is 128 as an autoregressive model.

\paragraph{Baselines}

TensorFlow~\cite{tensorflow} and PyTorch~\cite{pytorch} represent traditional ML frameworks that support both training and inference. We evaluated both of them on NVIDIA and AMD GPUs and PyTorch on Cambricon MLU. 
TorchScript~\cite{torchscript} is a library provided by PyTorch that converts PyTorch models into a more efficient, serialized format that can be used for inference and deployment. We evaluated TorchScript on NVIDIA and AMD GPUs.
TensorFlow-XLA (Accelerated Linear Algebra)~\cite{tensorflow_xla} is a domain-specific compiler for linear algebra used in TensorFlow. TensorFlow-XLA can automatically parallelize and vectorize TensorFlow computations to make use of the full power of modern hardware. We evaluated TensorFlow-XLA on NVIDIA GPU.
TensorRT~\cite{tensorrt} and MagicMind are deep learning inference optimizers developed by their vendors that enable high-performance inference based on high-performance kernels. We evaluated TensorRT on NVIDIA GPU and MagicMind on Cambricon MLU.
TVM~\cite{tvm} is a ML compiler using loop-based IR. Ansor~\cite{ansor} is an automated scheduling tool for TVM, representing the state-of-the-art for tensor compilers targeting computation kernels. We evaluated TVM/Ansor on NVIDIA and AMD GPUs.

\paragraph{Results}

The evaluation results, as presented in \Cref{fig:e2e}(a), demonstrate that \sys{} outperforms all baselines on A100. \sys{} achieves an average speed-up of $9.7\times$, $7.3\times$, $8.2\times$, and $4.1\times$ over PyTorch, TorchScript, TensorFlow, and TensorFlow-XLA, respectively. Additionally, compared to TensorRT, \sys{} accelerates models by $28\%$ on average and achieves a speed-up of $1.98\times$ on the GPT-2 model. Moreover, when compared to TVM, \sys{} achieves an average speed-up of $1.98\times$.

For BERT and ViT, \sys{} achieves similar performance as TensorRT due to GEMM taking up most of the execution time, thus limiting optimization opportunities. Meanwhile, for shuffleNet and EfficientNet, \sys{} achieves similar performance as TVM. This is because \sys{} spends more time on convolution operations compared to TVM, which affects the overall performance.

On MI100, \sys{} outperforms all baselines, achieving an average speed-up of $3.1\times$, $2.7\times$, and $18.0\times$ over PyTorch, TorchScript, and TensorFlow, respectively, as shown in \Cref{fig:e2e}(b). Compared to TVM, \sys{} achieves an average speed-up of $1.24\times$. The lower acceleration ratio on AMD GPU is due to the ratio of peak computation performance to memory bandwidth being lower than that of NVIDIA GPU. The memory access operation takes less time, resulting in a less significant optimization effect of \sys{}.

Finally, as demonstrated in \Cref{fig:e2e}(c), on Cambricon MLU-370, \sys{} achieves an average speed-up of $3.1\times$ over PyTorch and a speed-up of $2.3\times$ over MagicMind. 
This result indicates that \sys{} supports different architectures and has good cross-platform optimization capabilities.

\subsection{Case Study}
\label{eval:case}

\begin{figure}[t]
    \centering
    \includegraphics[width=0.45\textwidth]{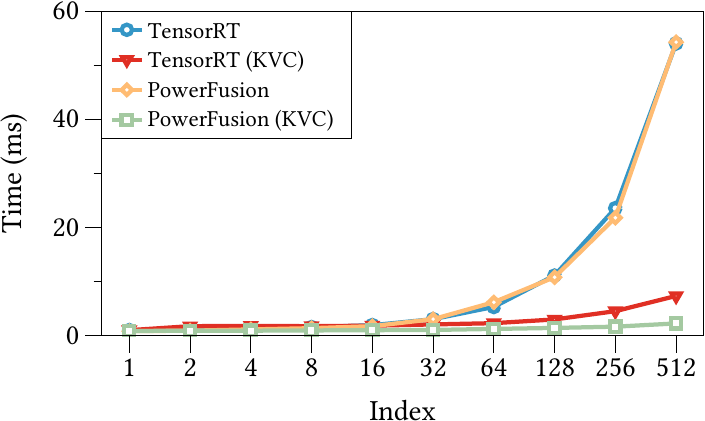}
    \caption{Performance of GPT-2 with different indices.}
    \label{fig:case}
\end{figure}

We use GPT-2 as an example to demonstrate how \sys{} outperforms existing frameworks. \Cref{fig:case} presents the performance of GPT-2 in a generation application on the A100 GPU. The figure shows that without utilizing KVCache, \sys{}'s performance is comparable to that of TensorRT. For this model, TensorRT manually optimizes the three fixed memory-intensive operators in the computation graph. In this case, \sys{} proposes similar optimization strategies compared to TensorRT, resulting in similar performance.


KVCache optimization is a widely deployed technique in generative models~\cite{kvcache}.
It leverages the characteristics of the GPT model by storing intermediate computation results in every iteration and reusing them later. 
To use the intermediate results from previous iterations, the KVCache optimization introduces two Concat operators in the attention layer, which concatenate the current and previous computation results. 
Thus, it reduces computation time by avoiding re-computation and maintains the equivalence of results.
As shown in \Cref{fig:case}, the execution time of TensorRT significantly decreases after applying the KVCache technique, especially for large input indices.

\begin{figure}[t]
    \centering
    \includegraphics[width=0.4\textwidth]{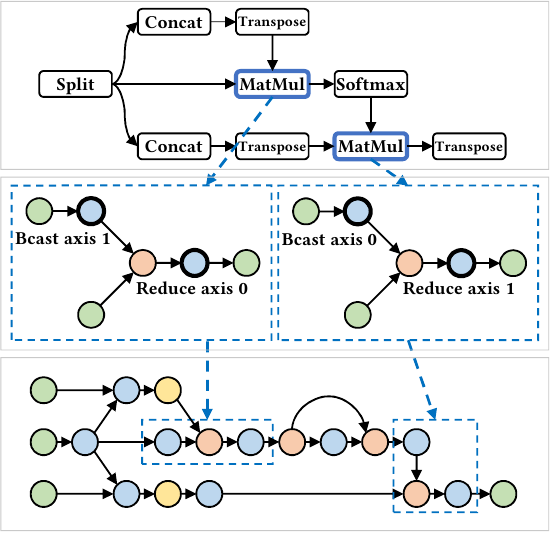}
    \caption{Case study of GPT2.}
    \label{fig:eval-case-graph}
\end{figure}

\sys{} can apply further optimization to GPT-2 after using KVCache. As each iteration only requires the computation of one token, the sequence length is always set to $1$. In the context of the attention layer in GPT-2 with KVCache, the two matrix multiplication operations degenerate into matrix-vector multiplication, which is limited by memory performance. Thus, using \sys{} to jointly optimize all operations of the attention layer and generate a memory-efficient kernel significantly improves program performance. \Cref{fig:eval-case-graph} demonstrates how \sys{} optimizes the attention layer, where the two matrix multiplications have identical shapes and computation patterns, but the program with the best performance after optimization uses distinct \ggraphs for them. These \ggraphs differ in their data slice pattern and axis of computation operation, making such optimizations challenging to achieve through loop-based approaches. Therefore, \sys{} exhibits a more significant performance advantage than TensorRT after applying KVCache, improving performance by up to $3.16\times$.

\subsection{Breakdown}

\begin{figure}[t]
    \centering
    \includegraphics[width=0.4\textwidth]{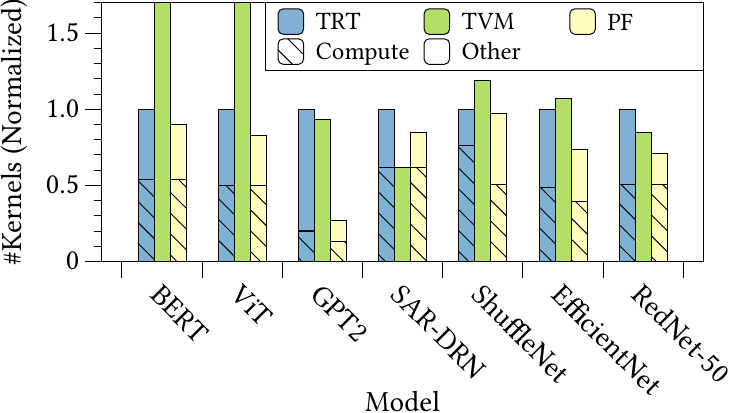}
    \caption{Breakdown of TensorRT, TVM, and \sys{}}
    \label{fig:eval-breakdown}
\end{figure}


To illustrate the source of \sys{}'s performance improvements, we break down the numbers of launched kernels of TensorRT, TVM, and \sys{}. 
As shown in \cref{fig:eval-breakdown}, we define computation operations like matrix multiplication and 2D-convolution in kernel libraries as computation kernels, while other operations are classified as memory kernels. Notably, since TVM regards all computation as tensor expressions and generates kernels for them, we only compare the total number of kernels to TVM.

The results indicate that for BERT, ViT, and RedNet50, \sys{} and TensorRT use the same number of computation kernels, but \sys{} generates fewer memory kernels, particularly in RedNet50, where the number of memory kernels is reduced by $59\%$ compared to TensorRT. On the other hand, TVM's total kernel numbers are higher than \sys{} and TensorRT. This observation demonstrates \sys{}'s stronger kernel fusion capability.

For SAR-DRN, TVM has the same number of kernels as \sys{} and TensorRT's computation kernel. This finding suggests that TVM can fuse all memory kernels in this model into computation kernels. However, \sys{} is limited by the vendor-provided kernel library and cannot fuse them into computation kernels. Our future work includes using \gir for computation-intensive kernels, which will give \sys{} the ability to fuse computation-intensive and memory-intensive kernels.

In contrast, for GPT-2, ShuffleNet, and EfficientNet, \sys{} reduces the number of computation kernels by combining them with memory kernel optimization. Consequently, the total number of kernels is reduced. For example, in GPT-2, the number of kernels generated by \sys{} is reduced by $73\%$ and $71\%$ compared to TensorRT and TVM, respectively. These findings demonstrate \sys{}'s strong optimization and code generation capabilities.

\subsection{Searching Time}

\begin{figure}[t]
    \centering
    \includegraphics[width=0.4\textwidth]{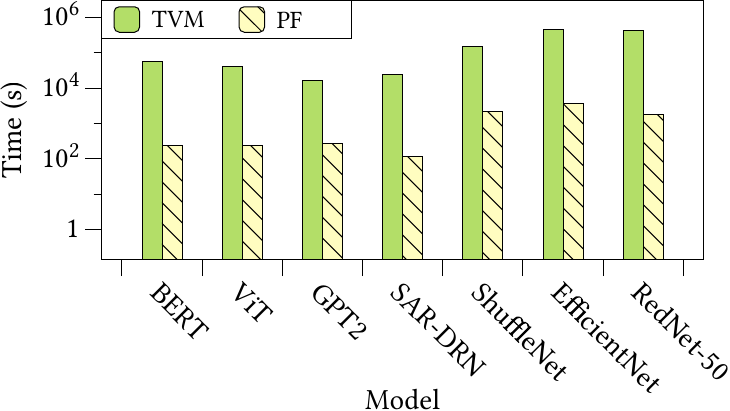}
    \caption{Searching time of TVM and \sys{}. }
    \label{fig:search-time}
\end{figure}

We compare the search times of \sys{} and TVM when optimizing various models. As depicted in \Cref{fig:search-time}, the search time for \sys{} is significantly shorter, at two orders of magnitude, compared to TVM. This outcome can be primarily attributed to two factors. First, our system directly uses the native library's implementation for computation-intensive operations, eliminating the need to search for it and thereby reducing search time. Second, we incorporate memory performance model pruning during the optimization process, which prevents the exploration of solutions involving substantial memory access and consequently decreases search overhead. Our evaluation results demonstrate that \sys{} can generate code for models with less time and offer significant advantages to users who require rapid model deployment.

\section{Related Works}
\label{ssec:related}



Many tensor compilers are capable of generating high-performance code for standalone deep learning operators, including TVM~\cite{tvm}, FlexTensor~\cite{flextensor}, Ansor~\cite{ansor}, AMOS~\cite{zheng2022amos}, Roller~\cite{zhu2022roller}, TensorIR~\cite{feng2022tensorir}, and Hidet~\cite{ding2023hidet}. However, their approaches do not model the data dependence between operators and cannot explore the data locality of nearby operators. Polyhedral approaches, including PPCG~\cite{ppcg} and PLUTO~\cite{pluto}, model the data dependence at the element level, which conducts too large searching space to find efficient solutions.

Other works accelerate DNN models at the graph level. TensorRT~\cite{tensorrt} and AITemplate~\cite{ai_template} manually optimize frequently-used patterns. TASO~\cite{jia2019taso}, Rammer~\cite{ma2020rammer}, and PET~\cite{pet} can combine existing operators to generate more efficient code. These pre-defined patterns and operator sets cannot cover the diversified memory access pattern in existing deep learning models. Astitch~\cite{astitch} and DNNFusion~\cite{dnnfusion} can generate memory-optimized code for unseen patterns, but their fusion is rule-based and overlooks many more efficient schedules.

Some domain-specific languages like Triton~\cite{tillet2019triton}, FreeTensor~\cite{tang2022freetensor}, and Graphene~\cite{hagedorn2023graphene} allows user to express tensor programs in finer granularity than operator-based computation graph but requires much more expert knowledge and human effort to explicitly define how the program should be executed.





\section{Conclusion}
\label{sec:conc}

In this paper, we proposed \sys{}, a compiler that optimizes memory performance through \gir by explicitly representing data movement patterns and fine-grained data dependence through instruction-level graph descriptions. By designing optimization strategies based on \gir and hardware abstraction, we are able to achieve significant speedup results on different hardware. Our evaluation results demonstrated that \sys{} achieves up to $1.97\times$, $2.93\times$, and $16.91\times$ speedup ($1.28\times$, $1.23\times$, and $2.31\times$ on average), respectively, compared to current most performant frameworks.

\bibliographystyle{plain}
\bibliography{refs}

\begin{thebibliography}{10}

\bibitem{magicmind}
{MagicMind}.
\newblock
  \url{https://https://www.cambricon.com/index.php?m=content&c=index&a=lists&catid=378}.

\bibitem{torchscript}
{TorchScript}.
\newblock \url{https://pytorch.org/docs/stable/jit.html}.

\bibitem{tensorflow_xla}
Xla: Optimizing compiler for tensorflow.
\newblock \url{https://www.tensorflow.org/xla}, 2017.

\bibitem{tensorflow}
Mart{\'\i}n Abadi, Paul Barham, Jianmin Chen, Zhifeng Chen, Andy Davis, Jeffrey
  Dean, Matthieu Devin, Sanjay Ghemawat, Geoffrey Irving, Michael Isard, et~al.
\newblock Tensorflow: A system for large-scale machine learning.
\newblock In {\em 12th USENIX symposium on operating systems design and
  implementation (OSDI 16)}, pages 265--283, 2016.

\bibitem{pluto}
Uday Bondhugula, Albert Hartono, J.~Ramanujam, and P.~Sadayappan.
\newblock A practical automatic polyhedral program optimization system.
\newblock In {\em ACM SIGPLAN Conference on Programming Language Design and
  Implementation (PLDI)}, June 2008.

\bibitem{tvm}
Tianqi Chen, Thierry Moreau, Ziheng Jiang, Haichen Shen, Eddie~Q. Yan, Leyuan
  Wang, Yuwei Hu, Luis Ceze, Carlos Guestrin, and Arvind Krishnamurthy.
\newblock {TVM:} end-to-end optimization stack for deep learning.
\newblock {\em CoRR}, abs/1802.04799, 2018.

\bibitem{bert}
Jacob Devlin, Ming{-}Wei Chang, Kenton Lee, and Kristina Toutanova.
\newblock {BERT:} pre-training of deep bidirectional transformers for language
  understanding.
\newblock {\em CoRR}, abs/1810.04805, 2018.

\bibitem{ding2023hidet}
Yaoyao Ding, Cody~Hao Yu, Bojian Zheng, Yizhi Liu, Yida Wang, and Gennady
  Pekhimenko.
\newblock Hidet: Task-mapping programming paradigm for deep learning tensor
  programs.
\newblock In {\em Proceedings of the 28th ACM International Conference on
  Architectural Support for Programming Languages and Operating Systems, Volume
  2}, pages 370--384, 2023.

\bibitem{vit}
Alexey Dosovitskiy, Lucas Beyer, Alexander Kolesnikov, Dirk Weissenborn,
  Xiaohua Zhai, Thomas Unterthiner, Mostafa Dehghani, Matthias Minderer, Georg
  Heigold, Sylvain Gelly, Jakob Uszkoreit, and Neil Houlsby.
\newblock An image is worth 16x16 words: Transformers for image recognition at
  scale, 2021.

\bibitem{feng2022tensorir}
Siyuan Feng, Bohan Hou, Hongyi Jin, Wuwei Lin, Junru Shao, Ruihang Lai, Zihao
  Ye, Lianmin Zheng, Cody~Hao Yu, Yong Yu, et~al.
\newblock Tensorir: An abstraction for automatic tensorized program
  optimization.
\newblock pages 804--817, 2023.

\bibitem{hagedorn2023graphene}
Bastian Hagedorn, Bin Fan, Hanfeng Chen, Cris Cecka, Michael Garland, and Vinod
  Grover.
\newblock Graphene: An ir for optimized tensor computations on gpus.
\newblock In {\em Proceedings of the 28th ACM International Conference on
  Architectural Support for Programming Languages and Operating Systems, Volume
  3}, pages 302--313, 2023.

\bibitem{jia2019taso}
Zhihao Jia, Oded Padon, James Thomas, Todd Warszawski, Matei Zaharia, and Alex
  Aiken.
\newblock Taso: optimizing deep learning computation with automatic generation
  of graph substitutions.
\newblock In {\em Proceedings of the 27th ACM Symposium on Operating Systems
  Principles}, pages 47--62, 2019.

\bibitem{li2021involution}
Duo Li, Jie Hu, Changhu Wang, Xiangtai Li, Qi~She, Lei Zhu, Tong Zhang, and
  Qifeng Chen.
\newblock Involution: Inverting the inherence of convolution for visual
  recognition.
\newblock In {\em Proceedings of the IEEE/CVF Conference on Computer Vision and
  Pattern Recognition}, pages 12321--12330, 2021.

\bibitem{ma2020rammer}
Lingxiao Ma, Zhiqiang Xie, Zhi Yang, Jilong Xue, Youshan Miao, Wei Cui,
  Wenxiang Hu, Fan Yang, Lintao Zhang, and Lidong Zhou.
\newblock Rammer: Enabling holistic deep learning compiler optimizations with
  rtasks.
\newblock In {\em Proceedings of the 14th USENIX Conference on Operating
  Systems Design and Implementation}, pages 881--897, 2020.

\bibitem{dnnfusion}
Wei Niu, Jiexiong Guan, Yanzhi Wang, Gagan Agrawal, and Bin Ren.
\newblock Dnnfusion: accelerating deep neural networks execution with advanced
  operator fusion.
\newblock In {\em Proceedings of the 42nd ACM SIGPLAN International Conference
  on Programming Language Design and Implementation}, pages 883--898, 2021.

\bibitem{kvcache}
Reiner Pope, Sholto Douglas, Aakanksha Chowdhery, Jacob Devlin, James Bradbury,
  Anselm Levskaya, Jonathan Heek, Kefan Xiao, Shivani Agrawal, and Jeff Dean.
\newblock Efficiently scaling transformer inference.
\newblock {\em CoRR}, abs/2211.05102, 2022.

\bibitem{pytorch}
{Tensors and Dynamic neural networks in Python with strong GPU acceleration.}
\newblock \url{https://pytorch.org}, 2017.

\bibitem{radford2019language}
Alec Radford, Jeff Wu, Rewon Child, David Luan, Dario Amodei, and Ilya
  Sutskever.
\newblock Language models are unsupervised multitask learners.
\newblock 2019.

\bibitem{halide}
Jonathan Ragan-Kelley, Connelly Barnes, Andrew Adams, Sylvain Paris, Fr{\'e}do
  Durand, and Saman Amarasinghe.
\newblock Halide: A language and compiler for optimizing parallelism, locality,
  and recomputation in image processing pipelines.
\newblock In {\em Proceedings of the 34th ACM SIGPLAN Conference on Programming
  Language Design and Implementation}, PLDI '13, 2013.

\bibitem{efficient_net}
Mingxing Tan and Quoc~V. Le.
\newblock Efficientnet: Rethinking model scaling for convolutional neural
  networks, 2020.

\bibitem{tang2022freetensor}
Shizhi Tang, Jidong Zhai, Haojie Wang, Lin Jiang, Liyan Zheng, Zhenhao Yuan,
  and Chen Zhang.
\newblock Freetensor: a free-form dsl with holistic optimizations for irregular
  tensor programs.
\newblock In {\em Proceedings of the 43rd ACM SIGPLAN International Conference
  on Programming Language Design and Implementation}, pages 872--887, 2022.

\bibitem{tensorrt}
{NVIDIA TensorRT}: Programmable inference accelerator.
\newblock \url{https://developer.nvidia.com/tensorrt}, 2017.

\bibitem{tillet2019triton}
Philippe Tillet, Hsiang-Tsung Kung, and David Cox.
\newblock Triton: an intermediate language and compiler for tiled neural
  network computations.
\newblock In {\em Proceedings of the 3rd ACM SIGPLAN International Workshop on
  Machine Learning and Programming Languages}, pages 10--19, 2019.

\bibitem{ppcg}
Sven Verdoolaege and Gerda Janssens.
\newblock {Scheduling for PPCG}.
\newblock {\em CW Reports}, 6 2017.

\bibitem{pet}
Haojie Wang, Jidong Zhai, Mingyu Gao, Zixuan Ma, Shizhi Tang, Liyan Zheng,
  Yuanzhi Li, Kaiyuan Rong, Yuanyong Chen, and Zhihao Jia.
\newblock Pet: Optimizing tensor programs with partially equivalent
  transformations and automated corrections.
\newblock In {\em 15th USENIX Symposium on Operating Systems Design and
  Implementation (OSDI 21)}, pages 37--54, 2021.

\bibitem{ai_template}
Bing Xu, Ying Zhang, Hao Lu, Yang Chen, Terry Chen, Mike Iovine, Mu-Chu Lee,
  and Zhijing Li.
\newblock {AITemplate}, October 2022.

\bibitem{sardrn}
Qiang Zhang, Qiangqiang Yuan, Jie Li, Zhen Yang, and Xiaoshuang Ma.
\newblock Learning a dilated residual network for sar image despeckling, 2018.

\bibitem{shuffle_net}
Xiangyu Zhang, Xinyu Zhou, Mengxiao Lin, and Jian Sun.
\newblock Shufflenet: An extremely efficient convolutional neural network for
  mobile devices.
\newblock In {\em 2018 {IEEE} Conference on Computer Vision and Pattern
  Recognition, {CVPR} 2018, Salt Lake City, UT, USA, June 18-22, 2018}, pages
  6848--6856. Computer Vision Foundation / {IEEE} Computer Society, 2018.

\bibitem{ansor}
Lianmin Zheng, Chengfan Jia, Minmin Sun, Zhao Wu, Cody~Hao Yu, Ameer Haj-Ali,
  Yida Wang, Jun Yang, Danyang Zhuo, Koushik Sen, et~al.
\newblock Ansor: generating high-performance tensor programs for deep learning.
\newblock In {\em 14th USENIX Symposium on Operating Systems Design and
  Implementation (OSDI 20)}, pages 863--879, 2020.

\bibitem{zheng2022amos}
Size Zheng, Renze Chen, Anjiang Wei, Yicheng Jin, Qin Han, Liqiang Lu, Bingyang
  Wu, Xiuhong Li, Shengen Yan, and Yun Liang.
\newblock Amos: enabling automatic mapping for tensor computations on spatial
  accelerators with hardware abstraction.
\newblock In {\em ISCA}, pages 874--887, 2022.

\bibitem{flextensor}
Size Zheng, Yun Liang, Shuo Wang, Renze Chen, and Kaiwen Sheng.
\newblock Flextensor: An automatic schedule exploration and optimization
  framework for tensor computation on heterogeneous system.
\newblock In {\em Proceedings of the Twenty-Fifth International Conference on
  Architectural Support for Programming Languages and Operating Systems}, pages
  859--873, 2020.

\bibitem{astitch}
Zhen Zheng, Xuanda Yang, Pengzhan Zhao, Guoping Long, Kai Zhu, Feiwen Zhu,
  Wenyi Zhao, Xiaoyong Liu, Jun Yang, Jidong Zhai, Shuaiwen~Leon Song, and Wei
  Lin.
\newblock Astitch: Enabling a new multi-dimensional optimization space for
  memory-intensive ml training and inference on modern simt architectures.
\newblock In {\em Proceedings of the 27th ACM International Conference on
  Architectural Support for Programming Languages and Operating Systems},
  ASPLOS '22, page 359–373, New York, NY, USA, 2022. Association for
  Computing Machinery.

\bibitem{zhu2022roller}
Hongyu Zhu, Ruofan Wu, Yijia Diao, Shanbin Ke, Haoyu Li, Chen Zhang, Jilong
  Xue, Lingxiao Ma, Yuqing Xia, Wei Cui, et~al.
\newblock $\{$ROLLER$\}$: Fast and efficient tensor compilation for deep
  learning.
\newblock In {\em 16th USENIX Symposium on Operating Systems Design and
  Implementation (OSDI 22)}, pages 233--248, 2022.

\end{thebibliography}
\end{document}